\documentclass{article}
\usepackage[margin=1.1in]{geometry}
\usepackage{authblk}
\usepackage[utf8]{inputenc}
\usepackage{amsmath}
\usepackage{amsfonts}
\usepackage{amsthm}
\usepackage{mathtools}
\usepackage{mathrsfs}
\usepackage{enumitem}
\usepackage{graphicx}
\usepackage{scrextend}
\usepackage{blindtext}
\usepackage{caption}
\usepackage{url}
\usepackage{subcaption}
\usepackage{circuitikz}
\usepackage{listings}
\usepackage{color}
 \usepackage{microtype}
\usepackage{graphicx}
\usepackage{bbm}
\usepackage{bm}
\usepackage{booktabs}
\usepackage{array,multirow}
\usepackage{comment}
\usepackage{parskip}
\newcommand{\specialcell}[2][c]{%
\begin{tabular}[#1]{@{}c@{}}#2\end{tabular}}
\def\BibTeX{{\rm B\kern-.05em{\sc i\kern-.025em b}\kern-.08em
    T\kern-.1667em\lower.7ex\hbox{E}\kern-.125emX}}
\usepackage{balance}
\newcolumntype{P}[1]{>{\centering\arraybackslash}p{#1}}
\usepackage{hyperref}
\usepackage[linesnumbered,lined,commentsnumbered,ruled]{algorithm2e}
\usepackage{amssymb}% http://ctan.org/pkg/amssymb
\usepackage{pifont}% http://ctan.org/pkg/pifont
\let\oldReturn\Return
\renewcommand{\Return}{\State\oldReturn}

\title{\textbf{Group-Feature (Sensor) Selection With Controlled Redundancy Using Neural Networks}}

\author[1]{Aytijhya Saha}
\author[1]{Nikhil R. Pal}

\affil[1]{Indian Statistical Institute, Kolkata, India.}
\affil[ ]{\textit {\{mb2308,nikhil\}@isical.ac.in}}
            
\begin{document}
\date{}

\maketitle
\begin{abstract}
In this work, we present a novel embedded feature selection method based on a Multi-layer Perceptron (MLP) network and generalize it for group-feature or sensor selection problems, which can control the level of redundancy among the selected features or groups and it is computationally more efficient than the existing ones in the literature. Additionally, we have generalized the group lasso penalty for feature selection to encompass a mechanism for selecting valuable groups of features while simultaneously maintaining control over redundancy. We establish the monotonicity and convergence of the proposed algorithm, with a smoothed version of the penalty terms, under suitable assumptions. The effectiveness of the proposed method for both feature selection and group feature selection is validated through experimental results on various benchmark datasets. The performance of the proposed methods is compared with some state-of-the-art methods. 
\end{abstract}


% \begin{highlights}
% \end{highlights}


\section{Introduction}
\label{intro}
Feature selection is a crucial dimension reduction
step with wide-ranging applications.  It can be conducted in supervised or unsupervised modes. In the case of supervised feature selection,  its primary objective is to reduce the dimension of the feature space maintaining the class discrimination power (for classification) and accuracy of prediction (for regression-type problems). On the other hand, for unsupervised approaches, feature selection tries to reduce the dimension preserving some task-independent attributes, such as the geometry of the original data in the lower dimensional space. Consequently, feature selection reduces the costs associated with measurement, system design, and decision-making. This paper concentrates on supervised feature selection tasks.

%Furthermore, different features may have different levels of importance to a specific application, some features may even exhibit a negative influence on a given task. Consequently, feature selection, which aims to identify and retain the most discriminative or informative features from the input data, has remained a prominent research focus for an extended period.

For a given problem, different features may play different roles, some may be more important than others, while some may be derogatory or irrelevant. Keeping this in mind, Chakraborty and Pal \cite{Chakraborty2008group}, have classified features into four groups: (i) essential features that must be selected to solve the given problem, if any of these features are removed, the problem cannot be solved successfully ; (ii) redundant features, which is a group of useful and dependent features, all of which may not be needed to solve the given problem; (iii) derogatory features are those features that must be rejected because such features can create problems during learning; and (iv) indifferent features, which neither help the learning nor create any problem to learning, but may increase the measurement cost.   
Since indifferent features make negligible / no contributions to the prediction process, they are thus considered candidates for removal. Although redundant features are useful, they exhibit dependency on one or more features, such as two correlated features. Therefore, to effectively solve the problem, we do not require all redundant features but only a subset of them. It is worth noting that completely removing redundancy from the selected features may not always be advisable. In such cases, if a measurement error occurs for a redundant feature, it could potentially disrupt the decision-making system's ability to function as intended.

In various applications, such as intelligent weld inspection and medical diagnostics, data from multiple sources/sensors are often utilized. For instance, in medical diagnostics, data from x-rays, CT images, and Magnetic Resonance images may be employed. Typically, from each sensory data multiple features are computed, and then those are fused to design the decision-making system, i.e., feature-level
sensor fusion for decision making  \cite{hall1992mathematical}. In such a situation, the importance of a sensor is determined by the group of features generated from the sensor output. 
So far, group feature selection (GFS) has been
successfully applied in several domains, such as gene finding \cite{meier2008group} and waveband selection \cite{subrahmanya2009sparse}.
 If we consider each group to contain only one feature, GFS reduces to the conventional feature selection problem. Thus, GFS represents a generalized feature selection problem.
Since the proposed method uses the multilayer perceptron (MLP) architecture with a minor variation, we briefly describe it for the sake of completeness. 

\subsection{{ Multilayer perceptron}}
{ Before proceeding further, we now introduce the Multilayer perception (MLP) network structure, as well as the notations and definitions that will be used throughout the paper. We summarize the notations used in Table \ref{notation}.}
%\subsection{Notations}

\begin{table}[!ht]
\caption{{ Different Notations Used} \label{notation}}
\begin{center}
{ \begin{tabular}{|ll|}\hline
$p$ &: Number of input features\\
$c$ &: Number of classes\\
$N$ &: Number of data points\\
$h$ &: Number of hidden nodes in an MLP network\\
$\mathbf{x}^i\in \mathbb{R}^p$ & : $i^{th}$ input vector\\
$\mathbf{y}^i\in \mathbb{R}^c$ & : Class label vector corresponding to $\mathbf{x}^i$\\
$\{\mathbf{x}^i,\mathbf{y}^i\}_{i=1}^N\subset \mathbb{R}^p\times\mathbb{R}^c $ &: Input-Output data set\\
$\mathbf{x}_{i}\in \mathbb{R}^N$ &: Vector formed by the $i$-th feature values over the \\ & training data\\
$\mathbf{Y}=\left(\mathbf{y}^{1}, \mathbf{y}^{2}, \ldots, \mathbf{y}^{N}\right)$ &: $c\times N$ matrix of class labels\\
$\mathbf{X}=\left(x_{i, j}\right)_{p \times N}=\left(\mathbf{x}^{1}, \mathbf{x}^{2}, \ldots, \mathbf{x}^{N}\right)=\left(\begin{array}{c}
\mathbf{x}_{1} \\
\mathbf{x}_{2} \\
\vdots \\
\mathbf{x}_{p}
\end{array}\right)
$ &: Input matrix of size $p\times N$\\
$v_{k i}$ &: Connecting weight between $i$ th input node and $k$ th \\ & hidden node\\
$\mathbf{v}_{i}=\left(v_{1 i}, v_{2 i}, \ldots, v_{h i}\right)^{T}$ for $i=1,2, \ldots, p$ &: Weight vector connecting $i$ th input node to all $h$ \\ & hidden nodes\\
$\mathbf{V}=\left(v_{k i}\right)_{h \times p}$ &: Weight matrix connecting the input and hidden layer\\
$\mathbf{U}=\left(u_{l k}\right)_{c \times h}$ &: The weight matrix connecting the hidden layer and \\ & output layer\\
$\mathbf{u}_{l}=\left(u_{l 1}, u_{l 2}, \ldots, u_{l h}\right)^{T}$ for $l=1,2, \ldots, c$&: The $l$-th row of the weight matrix $\mathbf{U}$\\
$\mathbf{w}=\left(\mathbf{u}_{1}, \ldots, \mathbf{u}_{c}, \mathbf{v}_{1}, \ldots, \mathbf{v}_{p}\right)^{T} \in \mathbb{R}^{h\times(p+c)}$ &: The matrix represents all weights $\mathbf{U}$ and $\mathbf{V}$ together\\
$f$ and $g$ &: Activation functions used at the hidden and  output \\ & layer nodes, respectively\\
%$G(\mathbf{z})=\left(g\left(z_{1}\right), \ldots, g\left(z_{h}\right)\right)^{T}$  & : $h$ dimensional output of the hidden layer\\
%$F(\mathbf{s})=\left(f\left(s_{1}\right), \ldots, f\left(s_{c}\right)\right)^{T}$ &: $c$ dimensional output of the output layer of the MLP\\ 
$\rho$ &: Pearson’s correlation coefficient\\\hline
\end{tabular}}
\end{center}
\end{table}

%Before proceeding, we first define some notation. 
We are given a dataset   $\{\mathbf{x}^i,\mathbf{y}^i\}_{i=1}^N\subset \mathbb{R}^p\times\mathbb{R}^c $, where $\mathbf{x}^i\in \mathbb{R}^p$ is the $i$ th input vector and $\mathbf{y}^i\in \mathbb{R}^c$ corresponds to its output ( here it is the class label vector). Let,
$$
\mathbf{X}=\left(x_{i, j}\right)_{p \times N}=\left(\mathbf{x}^{1}, \mathbf{x}^{2}, \ldots, \mathbf{x}^{N}\right)=\left(\begin{array}{c}
\mathbf{x}_{1} \\
\mathbf{x}_{2} \\
\vdots \\
\mathbf{x}_{p}
\end{array}\right)
$$
where $\mathbf{x}_{i}\in \mathbb{R}^N$, as explained in Table \ref{notation}, represents a vector consisting of the $i$-th feature values over the training data, and $\mathbf{Y}=\left(\mathbf{y}^{1}, \mathbf{y}^{2}, \ldots, \mathbf{y}^{N}\right)$. We consider a single-hidden layer multilayer perceptron
 network,  as used in \cite{wang2020feature}. The extension to multiple hidden layers is straightforward. Here, $N, p, h$, and $c$ denote the number
of data points, number of features (i.e, number of input-layer nodes), number of hidden-layer nodes, and number
of classes (i.e., number of output-layer nodes), respectively. 

The weight matrix connecting the input layer to the hidden layer is denoted as $\mathbf{V}=\left(v_{k i}\right)_{h \times p}$, where $v_{k i}$ represents the connecting weight between the $i$ th input node and the $k$ th $(k=1,2, \ldots, h)$ hidden node. Let $\mathbf{v}_{i}=\left(v_{1 i}, v_{2 i}, \ldots, v_{h i}\right)^{T}$ for $i=1,2, \ldots, p$ be the $i$ th column vector of $\mathbf{V}$ and 
$\mathbf{U}=\left(u_{l k}\right)_{c \times h}$ be the weight matrix connecting the hidden layer and the output layer. The $l$-th row of the weight matrix $\mathbf{U}$ is denoted by $\mathbf{u}_{l}=\left(u_{l 1}, u_{l 2}, \ldots, u_{l h}\right)^{T}$ for $l=1,2, \ldots, c$. For an easier representation, we combine  $\mathbf{U}$ and $\mathbf{V}$ into $\mathbf{w}=\left(\mathbf{u}_{1}, \ldots, \mathbf{u}_{c}, \mathbf{v}_{1}, \ldots, \mathbf{v}_{p}\right)^{T} \in \mathbb{R}^{h\times(p+c)}$. Let $f$ and $g$ be the activation functions used at the hidden and output layer nodes, respectively.

To facilitate further discussion, we introduce the following vector-valued functions:
\begin{align*}
   G(\mathbf{z})=\left(g\left(z_{1}\right), \ldots, g\left(z_{h}\right)\right)^{T} \quad \forall \mathbf{z}=\left(z_{1}, z_{2}, \ldots, z_{h}\right) \in \mathbb{R}^{h}\\
F(\mathbf{s})=\left(f\left(s_{1}\right), \ldots, f\left(s_{c}\right)\right)^{T} \quad \forall \mathbf{s}=\left(s_{1}, s_{2}, \ldots, s_{c}\right) \in \mathbb{R}^{c}.
\end{align*}

The empirical square loss function for the multilayer perceptron network is the following:
\begin{equation}
\label{e0}
E_0(\mathbf{w},\mathbf{X},\mathbf{Y})=\left\|F\left(\mathbf{U} G\left(\sum_{i=1}^{p} \mathbf{v}_{i} \mathbf{x}_{i}\right)\right)-\mathbf{Y}\right\|_{F}^{2}   
\end{equation}
where $\|.\|_F$ denotes the Frobenius norm.

Note that the weight vector $\mathbf{v}_{i}$ connects the $i$ th input node to all nodes in the first hidden layer. So, $\sum_{i=1}^{p} \mathbf{v}_{i} \mathbf{x}_{i}$ denotes the input matrix of the hidden layer and the output of the hidden layer denoted by $G(\cdot)$. Similarly, the input to the output layer is computed as $\mathbf{U} G(\cdot)$.  Finally, each output layer neuron applies an activation function. Thus, the outputs obtained from the output layer are $F(\mathbf{U} G(\cdot))$.

In the context of group-feature selection, we assume that the features are grouped into $s$ sensors/ feature-groups denoted as $G_1,\cdots,G_s$. Let $\mathbf{g}_i$ be the vector of weights that connect all the input nodes corresponding to $i$ th group-feature, $G_i$ to all the hidden nodes, and $n_i$ is the number of features in the group $G_i$. Thus, $\sum_{i=1}^s n_i=p$.

\subsection{{ Related Works on Feature Selection (FS)}}
The existing methods of feature selection are commonly categorized into three main types: filter \cite{liu1996probabilistic,dash2002feature,lazar2012survey,wang2022supervised}, wrapper \cite{kohavi1997wrappers}, and embedded/integrated approaches \cite{chakraborty2014feature,wang2020feature,zhang2019feature}. Filter methods evaluate the relevance of features via univariate statistics. The wrapper approach repeatedly uses a classifier on different subsets of features to search for the best subset of features for the given task. Embedded methods perform variable selection as a part of the learning procedure. The wrapper and embedded methods generally choose more useful features than those by filter methods. However, the evaluation mechanism of wrapper methods is quite time-consuming, especially for high-dimensional data. Embedded/integrated methods incorporate feature selection into the learning procedure. These methods typically yield more useful features compared to filter methods. Notably, embedded methods merge feature selection and learning tasks into a unified optimization procedure. As a result, they can capture subtle non-linear interactions between features and the learning tool, enhancing their performance.

In the literature, various feature selection approaches based on sparsity-inducing regularisation techniques have been presented \cite{7819557,jenatton2011structured,cong2016udsfs,Pang2023multitask}. The least absolute shrinkage and selection operator (Lasso)   \cite{tibshirani1996regression} is a very popular sparsity-based method of feature selection for linear multivariate regression. Recently, several works on feature selection using group lasso (GL) regularisation have been published \cite{zhang2019feature,wang2020feature,kang2021deterministic, wang2017convergence}, where authors have used GL penalty in the loss function of MLP network as follows:
\begin{equation}
\label{gl}
    GL=\sum_{i=1}^p \|\mathbf{v}_i\|_2=\sum_{i=1}^p\Big(\sum_{j=1}^h v_{ji}^2\Big)^\frac{1}{2}.
\end{equation}
Notations used here are the same as defined in the previous subsection.

The aforementioned methods excel at selecting useful features while eliminating derogatory and indifferent ones. But, they often struggle to control or monitor the use of dependent features. However, many real-life data sets such as gene expression data have numerous correlated/dependent useful features.  
 
 Pal and Malpani \cite{pal2012redundancy} proposed an integrated feature selection framework, using Pearson's correlation coefficient for penalizing the correlated features in radial basis function (RBF) networks. They added a penalty term $P_\mathbf{X}$ in \eqref{pal} to the RBF loss function, which is the following:
\begin{equation}
\label{pal}
    P_\mathbf{X}=\frac{1}{p(p-1)}\sum_{i=1}^p \gamma_i \sum_{j=1, j\neq i}^p \gamma_j \operatorname{dep}(\mathbf{x}_i,\mathbf{x}_j)
\end{equation}
where $\operatorname{dep}(\mathbf{x}_i,\mathbf{x}_j)$ is a measure of dependency between features $\mathbf{x}_i$ and $\mathbf{x}_j$ and $\gamma_i \in [0,1]$ is the $i^{th}$ feature modulator, which is realized using a modulator or gate function with a tuneable parameter. These $\gamma_i$s for all $i$ can be modeled using different modulating functions. One such choice 
is $\gamma_i=e^{-\beta_i^2}$ where $\beta_i$ is unrestricted.
Chakraborty and Pal \cite{chakraborty2014feature} proposed a general feature selection scheme using a multi-layer perception network that can control the level of redundancy in the set of selected features. Their method (named FSMLP-CoR) uses a penalty that has a similar form as equation \eqref{pal}. 

Chung et al. \cite{chung2017feature} used a fuzzy rule-based framework for feature selection with controlled redundancy (FRBS-FC). 
Banerjee and Pal \cite{banerjee2015unsupervised} use an unsupervised framework for feature selection with controlled redundancy (UFeSCoR). 

FSMLP-CoR, FRBS-FC, and UFeSCoR all use gate functions, one for each feature. Each gate function needs a tunable parameter thereby requiring additional parameters. To avoide the use of additional parameters, Wang et al. \cite{wang2020feature} offer an integrated scheme that operates directly on the neural network weights without the need for any extra parameters, using the following penalty: 

\begin{equation}
\label{wang}
    P=\frac{1}{p(p-1)}\sum_{i=1}^p  \|\mathbf{v}_i\|_2 \sum_{j=1, j\neq i}^p  \|\mathbf{v}_j\|_2 \operatorname{dep}(\mathbf{x}_i,\mathbf{x}_j).
\end{equation}

{ Similarly, our method operates directly on the network weights, without necessitating any extra parameters, as follows

\begin{equation}
\label{our}
    P=\frac{1}{hp(p-1)}\sum_{i=1}^p  \|\mathbf{v}_i\|_2 \sum_{j=1, j\neq i}^p  \operatorname{dep}(\mathbf{x}_i,\mathbf{x}_j).
\end{equation}

Notably, our approach stands out by its inherent naturalness and rationality, (which we elaborate on in Section \ref{fs}) and efficiency. Our penalty term for controlling redundancy is computationally more efficient than the existing ones. Our method avoids the need for computing $p(p-1)$ additional multiplications ($\|\mathbf{v}_j\|_2 \times \operatorname{dep}(\mathbf{x}_i,\mathbf{x}_j)$, for all $i,j=1,2,\cdots p$ and $i\neq j$), unlike previous approaches (See, e.g., equations \eqref{pal} and \eqref{wang}). This is a big advantage especially when the data dimension is high.}

\subsection{{ Related works on Group Feature Selection (GFS)}}
Over time, numerous research endeavors have explored the GFS problem, employing regularization techniques such as group lasso \cite{yuan2006model}, sparse group lasso \cite{simon2013sparse}, and Bayesian group lasso \cite{raman2009bayesian}. As mentioned earlier, neural networks have also been used to select groups of features.

GFS via group lasso has been used in several studies \cite{Pusponegoro2017group,du2016bayesian,tang2018group}. For example, Tang et al. \cite{tang2018group} formulated the GFS problem as a sparse learning problem in the framework of multiclass support vector machines using the multiclass group zero norm. They solved it by alternating direction method of multipliers (ADMM) in \cite{tang2018group}.  On the other hand, \cite{Pusponegoro2017group} used group lasso for selecting groups of features in the context of multivariate linear regression.

Chakraborty and Pal \cite{Chakraborty2008group} extended the feature selection method utilizing feature modulating gates to sensor selection or the selection of groups of features using both multilayer perceptron (MLP) and radial basis function (RBF) networks. In the case of an MLP, a single modulator is employed for each sensor or group of features.
For example, the modulator associated with the $l$ th sensor or group is defined as $\gamma_l=e^{-\beta_l^2}$.
And each feature of the $l^{th}$ group is multiplied by $\gamma_l$. Then the usual loss function of an MLP is used to train the network. There is no need to add any regularizer.
Each $\beta_l$ is initialized such that each $\gamma_l$ is nearly zero.

However, none of the above works considered controlling redundancy in the set of selected groups of features. 
It is worth noting that the concept of sensor modulating gate function has been used to select sensors with a control on the level of redundancy in the set selected sensors or groups of features.  For example, Chakraborty et al. \cite{chakraborty2014sensor} extended the group-feature selection method in \cite{Chakraborty2008group} to control the redundancy. They used the following regularizer in the loss function:

\begin{equation}
\label{Chakraborty2014group}
 P_\mathbf{X}=\frac{1}{s(s-1)}\sum_{i=1}^s \gamma_i \sum_{j=1, j\neq i}^s \gamma_j \operatorname{dep}(G_i,G_j)
\end {equation}

$\operatorname{dep}(G_i,G_j)$ is a measure of dependency between sensors $G_i$ and $G_j$ and $\gamma_i \in [0,1]$ is the $i^{th}$ group-feature modulator, similar to \eqref{pal}.

%In this work, we generalize the feature selection scheme of Zhang et al \cite{zhang2019feature} based on neural networks with the group lasso penalty in the context of the selection of groups of features.
In our subsequent discussion, we shall use the term sensors, feature groups, and groups of features interchangeably to represent the same thing.  We note here that features may be grouped based on the sensors that produce them or using some other criteria.

\subsection{Our contribution}
Below we summarize our primary contributions:
\begin{enumerate}
    \item We introduce an embedded feature selection method utilizing neural networks, capable of controlled removal of redundant features. Our formulation of the regularizer for redundancy control stands out from existing approaches, offering a more rational, computationally efficient, and effective solution.
    \item We extend the proposed feature selection method to accommodate the selection of sensors or groups of features.
  %  \item We have also extended the feature selection method based on neural networks with the group lasso penalty proposed by Zhang et al \cite{zhang2019feature}, in the context of sensor selection.
    \item Furthermore, we generalize the group lasso regularization and incorporate it alongside the penalty for controlling redundancy for sensor selection. In essence, we propose an integrated method for group feature (sensor) selection, that can adeptly remove derogatory groups of features, indifferent groups of features, and select useful groups of features with a control on the number of selected redundant/dependent groups of features.
    \item  We provide an analysis of the monotonicity and convergence
of the proposed algorithm, using a smoothed version of the regularizer, under some suitable assumptions. This analysis provides insights into the behavior and performance of our method, further validating its effectiveness and reliability.
\end{enumerate}

To the best of our knowledge, this is the first attempt to select groups of features or sensors, with controlled redundancy using the group lasso penalty in a neural framework, particularly using MLP networks.

The remaining part of the paper is structured as follows. Section \ref{sec:meth} discusses the methodologies of our work. Specifically, the feature selection scheme with redundancy control is described in Section \ref{fs}. Section \ref{gfs} extends our approach to group-feature selection by generalizing the methods outlined in Section \ref{fs}, along with incorporating techniques based on neural networks with the group lasso penalty \cite{zhang2019feature}. In Section \ref{theo}, the monotonicity and convergence
of the proposed algorithm, with a smoothed version of the penalty, are analyzed. Section \ref{sec:res} demonstrates compelling advantages of the proposed method through applications on real data sets. This
article is concluded in Section \ref{sec:conc}. 

\section{Methodology}
\label{sec:meth}
\noindent We first formulate a new loss function that will allow feature selection with controlled redundancy and then extend this framework to encompass group feature selection.
% A standard loss function for neural networks is,
% \begin{equation}
% \label{mse}
% E_0(\mathbf{w},\mathbf{X},\mathbf{Y})=\frac{1}{N}\sum_{i=1}^N \|\mathbf{o}_i-\mathbf{y}_i\|_2^2
% \end{equation}
% where $\mathbf{o}_i=(o_{i1},o_{i1},\cdots,o_{ip})^T$ is the output of the network. We assign $i$ th observation to $j$ th class, where $j=\arg \max_k o_{ik}$.

%is the output of the constructed neural network.

\subsection{Feature selection}
\label{fs}
\noindent We want to modify the loss function of our model to allow us to control the level of redundancy in the set of selected features during the feature selection process. Thus, the learning process should impose penalties on the selection of redundant features. As discussed in the previous section, we augment the loss in equation \eqref{e0} with a penalty that can discourage the use of 
numerous redundant features as follows:
 
\begin{equation}
    E=E_0(\mathbf{w},\mathbf{X},\mathbf{Y})+\lambda P(\mathbf{X},\mathbf{v}_1, \cdots, \mathbf{v}_j).
\end{equation}
To reduce the number of features, we need all weights connecting a derogatory or redundant feature to every node in the first hidden layer to have a very small magnitude, or almost zero. Hence, we consider the set of weights connected from a particular feature to the hidden layer nodes as a group. This motivates us to consider the following
\begin{equation}
\label{penalty-fs}
    P(\mathbf{X},\mathbf{v}_1, \cdots, \mathbf{v}_j)=\frac{1}{hp(p-1)}\sum_{i=1}^p \|\mathbf{v}_i\|_2\sum_{\substack{j=1\\j\neq i}}^p\operatorname{dep}(\mathbf{x}_i,\mathbf{x}_j).
\end{equation}
The parameter $\lambda\geq0$ is a regularizing constant that
governs the relative influence of the empirical error. The penalty term is designed to prevent selection of many dependent features. The measure of dependency between feature $i$ and feature $j$ is  computed using feature vectors $\mathbf{x}_i$ and $\mathbf{x}_j$  where the measure of dependency is denoted by $\operatorname{dep}(\mathbf{x}_i,\mathbf{x}_j)$. As explained earlier, $\mathbf{x}_i$ and $\mathbf{x}_j$  are the vectors with $N$ values of feature $i$ and feature $j$ respectively.  As a measure of dependency, we utilize the square of Pearson's correlation coefficient. The dependence measure $\operatorname{dep}(\mathbf{x}_i,\mathbf{x}_j)$ is fairly general.
It may, for example, be defined in terms of mutual information. { The factor $hp(p-1)$ helps to eliminate the dependency of the
regularizer on the number of features, $p$ and the number of hidden nodes, $h$. The rationale behind dividing by $p(p-1)$ stems from the fact that there are $p(p-1)$ terms of the form $||\mathbf{v}_i|| \operatorname{dep}(\mathbf{x}_i,\mathbf{x}_j)$ that are summed up. We also divide by $h$ because $||\mathbf{v}_i||$ represents the square root of the sum of squares of $h$ weights. This normalization technique is well-established in the literature, as evidenced by similar loss functions used by others (e.g., \cite{pal2012redundancy,chakraborty2014feature}).
From a mathematical standpoint, this factor may not hold significant implications, as we could achieve the same loss value by simply adjusting $\lambda$ proportionally. However, in practical terms, this normalization helps to eliminate the dependency of the parameter $\lambda$ on both the number of features and the number of hidden nodes.}

Note that the penalty for redundancy used in \cite{wang2020feature} is a bit different from the one we employ in our approach (see equation \eqref{wang}).
Their method penalizes the product $\|\mathbf{v}_i\|\|\mathbf{v}_j\|$, when $\operatorname{dep}(\mathbf{x}_i,\mathbf{x}_j)$ is high, which may potentially lead to confusion about whether we want to drop $\|\mathbf{v}_i\|$ or $\|\mathbf{v}_j\|$ or both. 
In contrast, our method penalizes $\|\mathbf{v}_i\|$ more, if the sum of the dependency values of $i$-th feature with all other features is large. Our approach aligns more intuitively with the goal of feature selection, as it emphasizes reducing the magnitude of $||\mathbf{v}_i||$ when it exhibits high inter-feature dependencies, making it a more appropriate choice. And also it avoids additional multiplications.

\subsection{Group feature (sensor) selection}
\label{gfs}
\noindent  
We begin by extending the feature selection method based on neural networks with the group lasso penalty proposed by Zhang et al. \cite{zhang2019feature} to the realm of \textit{group-feature} (or \textit{sensor}) selection. To effectively eliminate a ‘bad’ group of features,  the absolute value of every weight connecting the features of that group with all nodes in the hidden layer should be very small, or practically zero. Then, a natural extension of the group lasso penalty for feature selection \cite{zhang2019feature} (See equation \eqref{gl}) is 
\begin{equation}
\label{gl-ss}
  GL(\mathbf{g}_1, \cdots, \mathbf{g}_s)=\sum_{i=1}^s \frac{1}{n_ih} \|\mathbf{g}_i\|_2  
\end{equation}
where $\mathbf{g}_i$ is the vector of weights that connect all the input nodes corresponding to $i$ th group-feature to all the hidden nodes and $n_i$ is the number of features in the group $G_i$.
Unlike equation \eqref{gl}, the factor $n_ih$ is considered to make it independent of the number of features in each group and the number of hidden nodes. But, even after adding this $GL(\mathbf{g}_1, \cdots, \mathbf{g}_s)$ to the loss function, the model may select some relevant, but redundant groups of features, with high dependency on each other, which we do not want.

To address this concern, we first need to generalize the measure of redundancy/dependency between two groups (sensors). Assume feature $x$ is a desirable feature and feature $x'$ is highly related to (dependent on)  feature $x$.
By ``related/dependent", we mean that either of the two characteristics would suffice. Hence, the dependency between the two features is symmetric.  Now consider two groups of features $G$ and $G'$ such that for every $x \in G$, $\exists ~y \in G'$ where the dependency between $x$ and $y$ is very strong. In addition to those dependent features, $G'$ also includes some other features. In this situation, $G$ is highly linked with $G'$, but $G'$ is less dependent on $G$. Therefore, concerning group $G'$, $G$ is redundant, but the converse does not hold true.
Suppose, we have $s$ many groups of features, $G_1,G_2,\cdots, G_s$. When the dependency of $G_i$ on $G_j$, defined in equation \eqref{dep-ss}, is very high, we want the norm of $\mathbf{g}_i$ to be very small, or practically zero. Hence, a natural extension of the $P(\mathbf{\mathbf{X}})$ in equation \eqref{penalty-fs} is the following:
\begin{equation}
\label{ss-dep-penalty}
     P(\mathbf{X},\mathbf{g}_1, \cdots, \mathbf{g}_s)=\frac{1}{hs(s-1)}\sum_{i=1}^s \frac{1}{n_i}\|\mathbf{g}_i\|_2\sum_{\substack{j=1\\j\neq i}}^s\operatorname{dep}(G_i,G_j)
\end{equation}
where 
\begin{equation}
\label{dep-ss}
   \operatorname{dep}(G_i,G_j)=\underset {\mathbf{x}_l \in G_i} {\text{avg}}\underset{\mathbf{x}_m \in G_j}{\text{max}} \rho^2(\mathbf{x}_l,\mathbf{x}_m).
\end{equation}
Here, $\rho$ denotes Pearson's correlation coefficient.
A similar but distinct form of dependency, as expressed in the above equation, has been employed in the context of group-feature selection by Chakraborty and Pal \cite{chakraborty2014sensor}.  However, they incorporated additional parameters for sensor selection. Moreover, the penalty for redundancy, while related, differs from their approach.
It is worth noting that this dependency measure is asymmetric, i.e, $\operatorname{dep}(G_i,G_j) \neq \operatorname{dep}(G_j,G_i)$. Equation \eqref{dep-ss} is a plausible measure of group-dependency, but there could be alternative choices too, as suggested in \cite{chakraborty2014sensor}.

Finally, combining both the penalty terms (as expressed in equation \eqref{gl-ss}, \eqref{ss-dep-penalty}), the loss function that we consider is as follows:
\begin{equation}
\label{loss}   E(\mathbf{w},\mathbf{X},\mathbf{Y})=E_0(\mathbf{w},\mathbf{X},\mathbf{Y}) + \lambda P(\mathbf{X},\mathbf{g}_1, \cdots, \mathbf{g}_s) + \mu GL(\mathbf{g}_1, \cdots, \mathbf{g}_s).
\end{equation}
Here, $\lambda$ and $\mu$ are the regularizing constants determining the severity of the respective penalty terms.

Using gradient descent, the weight vectors are updated as follows
%$$ 

\begin{align}
\label{deriv-u}
\mathbf{u}_i^{m+1} & =\mathbf{u}_i^m-\eta \frac{\partial}{\partial\mathbf{u}_i^m}E(\mathbf{w}_m,\mathbf{X},\mathbf{Y}) =\mathbf{u}_i^m-\eta \frac{\partial}{\partial\mathbf{u}_i^m}E_0(\mathbf{w}_m,\mathbf{X},\mathbf{Y}) \\
\nonumber
\mathbf{v}_j^{m+1} & =\mathbf{v}_j^m-\eta \frac{\partial}{\partial\mathbf{v}_j^m}E(\mathbf{w}_m,\mathbf{X},\mathbf{Y}) \\
\label{deriv-v}&=\mathbf{v}_j^m-\eta \frac{\partial}{\partial\mathbf{v}_j^m}E_0(\mathbf{w}_m,\mathbf{X},\mathbf{Y})-\frac{\eta\lambda}{hs(s-1)} \frac{\mathbf{v}_j^m}{n_i\|\mathbf{g}_k^m\|_2}\sum_{\substack{j=1\\j\neq k}}^s\operatorname{dep}(G_k,G_j)- \frac{1}{n_ih} \frac{\mathbf{v}_j^m}{\|\mathbf{g}^m_k\|_2},  
\end{align}
%$$
where $j$ th feature lies in $k$-th group $G_k$.
%$$
\begin{equation} \label{updateu}
\begin{aligned}
\frac{\partial E_0}{\partial\mathbf{u}_i^m}
= & \sum_{k=1}^N\left(f\left(\mathbf{u}_i^m G\left(\sum_{j=1}^p (\mathbf{v}_j^{m})^T  \mathbf{x}_j^k\right)\right)-y_i^k\right) \times f^{\prime}\left(\mathbf{u}_i^m G\left(\sum_{j=1}^p (\mathbf{v}_j^m)^T\mathbf{x}_j^k\right)\right)\left(G\left(\sum_{j=1}^p (\mathbf{v}_j^m)^T \mathbf{x}_j^k\right)\right)^T, \\
\end{aligned}
\end{equation}
%$$
\begin{equation}\label{updatev}
   \frac{\partial E_0}{\partial\mathbf{v}_j^m}
= \sum_{k=1}^N\sum_{i=1}^c\left(f\left(\mathbf{u}_i^m G\left(\sum_{j=1}^p (\mathbf{v}_j^m)^T \mathbf{x}_j^k\right)\right)-y_i^k\right) 
 \times f^{\prime}\left(\mathbf{u}_i^m G\left(\sum_{j=1}^p (\mathbf{v}_j^m)^T \mathbf{x}_j^k\right)\right) 
 \times \mathbf{u}_i^m \circ\left(G\left(\sum_{j=1}^p (\mathbf{v}_j^m)^T \mathbf{x}_j^k\right)\right)^{\prime} \mathbf{x}_j^k
\end{equation}
the symbol $\circ$ is the Hadamard product and $$\left(G\left(\sum_{j=1}^p (\mathbf{v}_j^m)^T \mathbf{x}_j^k\right)\right)^{\prime}=\left(g^{\prime}\left(\sum_{j=1}^p v_{1 j}^m \mathbf{x}_j^k\right), \ldots, g^{\prime}\left(\sum_{j=1}^p v_{h j}^m \mathbf{x}_j^k\right)\right),  m \in \mathbb{N}, j=1,2, \ldots, p; k=1,2, \ldots, N.$$
\subsection{Computational Complexity}
The scalability of the algorithm depends on the complexity of the update equations \eqref{deriv-u}, and \eqref{deriv-v}.
 The complexity for computing both $\frac{\partial E_0}{\partial\mathbf{u}_i^m}$ and $\frac{\partial E_0}{\partial\mathbf{v}_j^m}$ is $O(Np)$, for each $i=1,\cdots,c$ and $j=1,\cdots,p$ and for the last two terms in \eqref{deriv-v} is $O(n_k+s)$, for $k=1,\cdots,s$. Hence, the computational cost in each step is $O(Np^2+s+s^2)$, which is same as $O(Np^2)$. Separately, the cost for computing all $\operatorname{dep}(G_i,G_j)$ is $O(s^2)$, which is at most $O(p^2)$.
Consequently, the overall complexity is $O(Np^2)$.

\section{Theoretical properties}
\label{theo}
At the origin, the two penalty terms, $P()$ and $GL()$ of the loss function in equation \eqref{loss} are not differentiable. In order to make the loss function differentiable and enable the use of gradient descent optimization techniques, it becomes necessary to use a smoothing approximation approach. This involves introducing a differentiable smoothing function, denoted as $H$, so that $H(\mathbf{g}_i)$ serves as an approximation to $\|\mathbf{g}_i\|$.
The use of the smoothing technique to handle the non-differentiable penalty terms to facilitate the theoretical analysis is a well-established practice in the literature (\cite{wang2020feature,zhang2019feature,wang2017novel,chen2010smoothing}). The total loss function equation \eqref{loss} is then modified as
\begin{equation}
 \label{eq:loss-sm}
    E(\mathbf{w},\mathbf{X},\mathbf{Y})= E_0(\mathbf{w},\mathbf{X},\mathbf{Y}) + \lambda\frac{1}{hs(s-1)}\sum_{i=1}^s \frac{1}{n_i}H(\mathbf{g}_i)\sum_{\substack{j=1\\j\neq i}}^s\operatorname{dep}(G_i,G_j) +\mu\sum_{i=1}^p H(\mathbf{g}_i) 
\end{equation}
Thus, using  gradient descent, the update equation   at the $m$ th step is as follows:
\begin{equation}
\label{eq:update}
    \mathbf{w}^{m+1}=\mathbf{w}^{m}-\eta\frac{\partial E}{\partial\mathbf{w}^{m}}, \quad m \in \mathbb{N}.
\end{equation}
For analyzing the algorithm using smoothing group lasso penalty and redundancy control, let us first consider the four assumptions listed below, which align with the assumptions made in previous works \cite{wang2020feature,zhang2019feature}:
\begin{itemize}
    \item[A1:] The activation functions $f$ and $g$ are continuous differentiable on $\mathbb{R}$, and $f, g, f^{\prime}$ and $g^{\prime}$ are uniformly bounded on $\mathbb{R}$, i.e, there exists a constant $C>0$ such that
    
    $\sup _{x \in \mathbb{R}}\left\{|f(x)|,|g(x)|,\left|f^{\prime}(x)\right|,\left|g^{\prime}(x)\right|\right\} \leq C$.
    \item[A2:] The learning rate $\eta>0$ and satisfies that $\eta<\frac{1}{C_{8}}$, where, $C_{8}$ is as mentioned in \cite{wang2020feature}.
    \item[A3:] The weight sequence $\left\{\mathbf{w}^{m}\right\}_{m \in \mathbb{N}}$ is uniformly bounded in $\mathbb{R}^{h \times(p+c)}$.
    \item[A4:]  The stationary points of equation \eqref{eq:loss-sm} are at most countably infinite.
\end{itemize}
Then the following two theorems \cite{wang2020feature,zhang2019feature} are valid for our group-feature (sensor) selection method as well.\\\\
\textbf{Theorem 1 :} For the cost function defined by equation \eqref{eq:loss-sm}, and the weight sequence $\left\{\mathbf{w}^{m}\right\}_{m \in \mathbb{N}}$, generated by the iterative updating formula \eqref{eq:update}, suppose that the assumptions $\mathrm{A} 1-\mathrm{A} 3$ are valid, then 
$$
E\left(\mathbf{w}^{m+1},\mathbf{X},\mathbf{Y}\right) \leq E\left(\mathbf{w}^{m},\mathbf{X},\mathbf{Y}\right), \quad m=0,1, \ldots
$$
\textbf{Theorem 2 :} Suppose that the assumptions $\mathrm{A} 1-\mathrm{A} 3$ are satisfied. Then the weight sequence $\left\{\mathbf{w}^{m}\right\}_{m \in \mathbb{N}}$, generated by equation \eqref{eq:update} satisfies the following weak convergence:
$$
\lim _{m \rightarrow \infty}\left\|\frac{\partial E\left(\mathbf{w}^{m},\mathbf{X},\mathbf{Y}\right)}{\partial\mathbf{w}^{m}}\right\|=0.
$$
In addition, if the assumption A4 is valid, then the strong convergence also holds, i.e, $\exists \mathbf{w}^{*} \in \mathbb{R}^{h \times(p+c)}$ such that
$$
\lim _{m \rightarrow \infty} \mathbf{w}^{m}=\mathbf{w}^{*}.
$$
\textbf{Remark:} Theorem 1 and Theorem 2 deal with monotonicity and convergence respectively. It is worth noting that the equation \eqref{eq:loss-sm} can be rewritten as
\begin{align*}
     E = E_0(\mathbf{w},\mathbf{X},\mathbf{Y}) + \sum_{i=1}^p \lambda_i H(\mathbf{g}_i)
\end{align*}
where $\lambda_i = \mu + \frac{\lambda}{n_ihs(s-1)}\sum_{\substack{j=1\\j\neq i}}^s\operatorname{dep}(G_i,G_j)~;~i=1,\cdots p.$
Clearly, $\lambda_i$ s are bounded quantities. Hence, the proofs of these two theorems can be done in the same manner
as the proofs in \cite{wang2020feature}.

Although we have used smoothing to provide the theoretical guarantees, we did not use smoothing in the experimental studies. This choice is based on the observation that even if we do not use smoothing, the plots depicting the norm of weights vs. time are smooth enough for the datasets we have used (for example, see Figs. \ref{fig:iris},\ref{fig:landsat}). Consequently, the results with the non-smoothing penalty terms in equation \eqref{gl-ss} and \eqref{ss-dep-penalty} do not cause any issues for the practical use of our proposed method.

\section{Experimental results}
\label{sec:res}
In this section, we evaluate the performance of the proposed methods using real-world datasets and compare them with state-of-the-art methods.

Our experiment has three parts. First, we implement the \textit{feature selection} method proposed in Section \ref{fs} with a case study on six datasets, Iris, WBC, Sonar, Thyroid, SRBCT, and Leukemia. Second, we assess the performance of the \textit{group-feature selection} method proposed in Section \ref{gfs}. We compared our method with a few recent ensemble classifiers using four datasets- Lung \cite{li2017feature}, DLBCL \cite{shipp2002diffuse}, NCI60 \cite{james2013introduction} and Semeion in Section \ref{non-nn-expt}. We employed six datasets, namely, Iris, Iris-2, Gas Sensor, LRS, Smartphone, and LandSat data for this part of our analysis. {  Lastly, in subsection \ref{subsec-rna}, we have shown an application on RNA sequence data (RNA-Seq 1 and RNA-Seq 2) and compared our result with those of Tang et al. \cite{tang2018group}.} Table \ref{tab:data} summarizes all datasets. The datasets can be found online. \footnote{SRBCT and Leukemia datasets are available at \url{https://file.biolab.si/biolab/supp/bi-cancer/projections/}. All other datasets are collected from UCI Machine Learning Repository (\url{https://archive.ics.uci.edu/)}. Note that RNA-Seq 2 is a manual modification of RNA-Seq 1.}.

%every dataset as follows: $x^{\prime}=(x-\mu)/\sigma$, where $x^{\prime}$ is the normalized feature value, and $\mu$ and $\sigma$ are the mean and standard deviation of the feature $x$,respectively.

\begin{table}[!htb]
    \centering
    \caption{Description of the Datasets\label{tab:data}}
    \resizebox{0.54\linewidth}{!}{
    \begin{tabular}{lccc}
    \toprule
      \textbf{Dataset}   & \textbf{Dataset Size} & \textbf{Features} &  \textbf{Classes}\\
      \midrule
       Iris  &  150 & 4 & 3  \\
       Thyroid & 215 & 5 & 3\\
       Iris 2  &  150 & 7 & 3 \\
       WBC & 699 & 9 & 2\\
       LandSat & 6435 & 44 & 6\\
Sonar & 208 & 60 & 2\\
LRS  & 531 &   93 & 10\\
Gas sensor  & 13,790  & 128 & 6\\
Semeion & 1593 & 257 & 10\\
Smartphones & 10,299 & 561 & 6\\ 
SRBCT & 83 & 2,308 & 4\\
Lung & 203 & 3,313 & 2\\
Leukemia & 72 & 5,147 & 2\\
DLBCL & 77 & 5,470 & 2\\
NCI60 & 64 & 6,831 & 2\\
{RNA Seq 1} & { 801} & { 20,531} & { 5}\\
RNA Seq 2 & { 801} & { 20,530} & { 5}\\

\bottomrule
    \end{tabular}}
\end{table}

We employ multi-layer perceptron networks with one hidden layer for our experiments. However, depending on the requirements, multiple hidden layers can also be utilized. Additionally, we utilize the sigmoid activation function for both the hidden and output layers.
The training process is conducted using the gradient descent method, with a maximum of 500 iterations set for all experiments.

Before using any data set, we do Z-score normalization of every feature of the data set.
We have adopted the scheme in Algorithm 1 for both feature selection (FS) and Group-feature selection (GFS). In Step 2 of Algorithm 1, the data set is first  randomly split into a training set ($80\%)$ and test set ($20\%$), except for the LandSat data. For the LandSat data we have
used $4435$ instances in the training set and $2000$ instances
to test the classification accuracy in accordance with the UCI repository and prior works (Chakraborty and Pal \cite{chakraborty2014sensor}). 
The training set and test set generated in Step 2 are denoted by $\mathbf{X}_{\text{TR}}$ and   $\mathbf{X}_{\text{TE}}=\mathbf{X}-\mathbf{X}_{\text{TR}}$, respectively. Then, Step 3 to Step 10 use a 10-fold cross-validation procedure on the training data $\mathbf{X}_{\text{TR}}$ selected in Step 2  to determine the desirable number of hidden nodes. {  To achieve this, $\mathbf{X}_{\text{TR}}$ is randomly split into ten folds, of which nine folds are used to train a number of  MLPs each with a different number of hidden nodes. The MLP is trained with the loss function  in equation \eqref{e0}. Each trained network is then validated on the remaining fold.
This is repeated for 10 times so that each of the 10 folds is used as a validation set. For the training set $\mathbf{X}_{\text{TR}}$  generated in Step 2, the desirable number of hidden nodes is computed in Step 10, which selects the number of hidden nodes for which the average validation error in the inner loop of Algorithm 1 (Steps 4 - 9) is the minimum.  After choosing the architecture, an MLP is trained on the entire $\mathbf{X}_{\text{TR}}$  generated in Step 2 using the loss in \eqref{loss}.  
Then, in Step 12, we select the features (group features) with weight vectors having norms greater than or equal to  $\theta=0.1\times \max_i \|\mathbf{v}_i\|_2$ ($\theta=0.1\times \max_i \|\mathbf{g}_i\|_2$) to obtain the reduced training and test data, as done in \cite{chakraborty2014feature}. 
Finally, in Step 13 of Algorithm 1, we train an MLP network minimizing the loss in \eqref{e0} using the lower dimensional version of $\mathbf{X}_{\text{TR}}$ defined using the selected features. This network is then tested on the lower dimensional version of $\mathbf{X}_{\text{TE}}$ 
 in Step 13 of the algorithm.} The entire procedure is repeated 10 times independently and the average test accuracy, denoted by FinaTestAcc, computed  in Step 16 of the Algorithm 1 is reported.
 
{ 
In the case of supervised feature selection, the   most popular index of performance evaluation is average accuracy, where the average is computed over several repeats of a two-level cross validation experiment. In some studies, authors also consider the average number of features selected by the method. Further, to demonstrate the consistency of a feature selection method, sometimes number of distinct features selected is also used as an evaluation index.   Moreover, since our method can  control the level of redundancy in the set of selected features, we have also considered the average  absolute correlation as well as the maximum absolute correlation among the selected features as performance indices. While comparing with other methods, we could use  only those indices that are used by the comparing methods.} 

\begin{comment}
 \begin{algorithm}
\label{algo:1}
\caption{Algorithm for FS(GFS)}\label{euclid}
\begin{algorithmic}[1]
\Require $\mu, \lambda\geq 0$, data $\mathbf{X}=\{\mathbf{x}^i,\mathbf{y}^i\}_{i=1}^N$
\For{i = 1 to 10} 
\State Randomly select $80\%$ distinct samples of $\mathbf{X}$ to
obtain $\mathbf{X}_{\text{TR}}$ and set  $\mathbf{X}_{\text{TE}}=\mathbf{X}-\mathbf{X}_{\text{TR}}$
\State Divide the data $\mathbf{X}_{\text{TR}}$ into 10 equal parts $\mathbf{X}_{\text{TR}}^1,\cdots,\mathbf{X}_{\text{TR}}^{10}$
\For{i = 2 to 20}
 \For{j = 1 to 10}
 \State Train an MLP network with $i$ hidden nodes and empirical loss  equation \eqref{e0} on $\bigcup_{i\neq j}\mathbf{X}_{\text{TR}}^i$ 
\State        Test the trained network on $\mathbf{X}_{\text{TR}}^k$ and let $e(i,j)$ be the test error
  \EndFor  
  \EndFor  
 \State  h = $ \arg \min_i \{\frac{1}{10}\sum_{j=1}^{10} e(i,j)\}$.

\State Train an MLP network with $h$ hidden nodes and loss  equation \eqref{loss} on $\mathbf{X}_{\text{TR}}$ 
\State Select features (groups) using threshold $\theta=0.1\times \max_i \|\mathbf{v}_i\|$ to obtain reduced data, $\mathbf{X}_{\text{TR}}^{'}$ and $\mathbf{X}_{\text{TE}}^{'}$.
\For{j = 1 to 10} 
 \State Train an MLP network with $h$ hidden nodes and empirical loss equation \eqref{e0} on $\mathbf{X}^{'}_{\text{TR}}$
 \State Calculate $a_j=$ test accuracy of the model on $\mathbf{X}^{'}_{\text{TR}}$
\EndFor  
\EndFor  
\Return $\text{FinalTestAcc}=\frac{1}{10}\sum_{j=1}^{10}a_j$
\end{algorithmic}
\end{algorithm}
\end{comment}

\begin{algorithm}
\label{algo:1}

\SetKwData{Left}{left}\SetKwData{This}{this}\SetKwData{Up}{up}
\SetKwFunction{Union}{Union}\SetKwFunction{FindCompress}{FindCompress}
\SetKwInOut{Input}{Input}\SetKwInOut{Output}{output}
%\DontPrintSemicolon
\SetAlgoLined
%\begin{algorithmic}[1]
\Input{$\mu, \lambda\geq 0$, data $\mathbf{X}=\{\mathbf{x}^i,\mathbf{y}^i\}_{i=1}^N$}
\BlankLine
\For{m = 1 to 10}{
Randomly select $80\%$ distinct samples of $\mathbf{X}$ to
obtain $\mathbf{X}_{\text{TR}}$ and set  $\mathbf{X}_{\text{TE}}=\mathbf{X}-\mathbf{X}_{\text{TR}}$\;
Divide the data  $\mathbf{X}_{\text{TR}}$ randomly into 10 equal parts $\mathbf{X}_{\text{TR}}^1,\cdots,\mathbf{X}_{\text{TR}}^{10}$\;
\For{i = 2 to 20}{
 \For{j = 1 to 10}{
Train an MLP network with $i$ hidden nodes and empirical loss   \eqref{e0} on $\bigcup_{i\neq j}\mathbf{X}_{\text{TR}}^i$\;
 Test the trained model on $\mathbf{X}_{\text{TR}}^k$ and let $e(i,j)$ be the test error
  }}
  h = $ \arg \min_i \{\frac{1}{10}\sum_{j=1}^{10} e(i,j)\}$\;
 Train an MLP network with $h$ hidden nodes and loss as defined in \eqref{loss} on $\mathbf{X}_{\text{TR}}$\; 
Select features (groups of features) using the threshold $\theta=0.1\times \max_i \|\mathbf{v}_i\|_2$ ($\theta=0.1\times \max_i \|\mathbf{g}_i\|_2$) to obtain the reduced data, $\mathbf{X}_{\text{TR}}^{'}$ and $\mathbf{X}_{\text{TE}}^{'}$\;
Train an MLP network with $h$ hidden nodes and empirical loss  \eqref{e0} on $\mathbf{X}^{'}_{\text{TR}}$\;
 Compute $a_m=$ test accuracy of that model on $\mathbf{X}^{'}_{\text{TE}}$\;
}
$\text{FinalTestAcc}=\frac{1}{10}\sum_{m=1}^{10}a_j$\;
%\end{algorithmic}
\caption{Algorithm for FS(GFS)}
\end{algorithm}

\begin{table}[!htb]
    \centering
    \caption{Feature selection results, setting $\mu=0$}
\label{tab:fsresult}
    \resizebox{0.69\linewidth}{!}{
    \begin{tabular}{lcccccc}
    \toprule
    \addlinespace
       \textbf{Dataset} & $\bm{\lambda}$  & 
       \specialcell{\textbf{Test}\\\textbf{Acc}} &
       \specialcell{\textbf{Distinct}\\\textbf{\# features}} &
        \specialcell{\textbf{Average}\\\textbf{\# features}} &
\specialcell{\textbf{Max abs}\\\textbf{corr}} &
\specialcell{\textbf{Avg abs}\\\textbf{corr}} \\ 
        \midrule
        IRIS  & 0 & 96.00 & 4 & 4 & 0.5898 & 0.9628 \\ 
        ~ & 10  & 95.03 & 2 & 2 & 0.4205 &	0.4205\\ 
        ~ & 20  & 94.07 & 2 & 1.5 & 0.4205 & 0.4205\\ \addlinespace
    \hline
        \addlinespace
        WBC & 0   & 94.04 & 9 & 9 & 0.8271 & 0.3831 \\
        ~ & 20  & 94.44 & 8 & 7.7 & 0.7627 & 0.3714 \\ 
        ~ & 50  & 90.61 & 8 & 6.4 & 0.6698 & 0.3724 \\ 
        \addlinespace
        \hline
        \addlinespace
        Thyroid & 0 &  96.12 & 5 & 5 & 0.7187 & 0.4135  \\ 
        ~ & 10  & 96.23 & 5 & 4.7 & 0.7187 & 0.4135 \\ 
        ~ & 20  & 94.37 & 5 & 3.8 & 0.6523 & 0.4405\\
        \addlinespace
        \hline
        \addlinespace
        SONAR & 0 & 82.50 & 60 & 60 & 0.8601 & 0.0828 \\ 
        ~ & 20  & 83.77 & 50 & 31.5 & 0.8080 & 0.0726 \\ 
        ~ & 50  & 84.52 & 46 & 30 & 0.8070 & 0.0699 \\ \addlinespace
        \hline
        \addlinespace
        SRBCT  & 0 & 86.02 & 2308 & 2308 & 0.9729 & 0.1572\\ 
        ~ & 20  & 92.55 & 1996 & 1209 &  0.9721 &	0.1448\\ 
        ~ & 50  & 98.00 & 784 & 437.4 & 0.9434	& 0.1492 \\
        \addlinespace
        \hline
        \addlinespace
        Leukemia  & 0 & 88.80
 & 5147
 & 5147
 & 0.9954 & 0.1702
\\ 
        ~ & 20  & 95.01 & 1498
 & 695.2
 &  0.9927 &	0.1383
  \\ 
        ~ & 50  & 96.13 & 704 & 299 & 0.9881 & 0.1475
 \\
 \bottomrule
    \end{tabular}}
\end{table}

\subsection{Feature selection results}
\label{subsec:fs-res}

In this subsection, we assess the performance of our proposed regularizer for redundancy control and compare it with other state-of-the-art methods. It is important to clarify that, we did not employ the group lasso penalty for feature selection, as this methodology is well-established in the existing literature \cite{zhang2019feature,wang2020feature}. In the realm of feature selection, our contribution lies in the formulation of the penalty for redundancy control, as expressed in equation \eqref{penalty-fs}. Thus, we present experimental results utilizing our proposed redundancy control regularizer, showcasing its effectiveness in the context of feature selection. However, in Subsection \ref{subsec:ss-res}, we incorporate the group lasso penalty alongside our proposed regularizer to control redundancy for group-feature selection within a neural network framework. To the best of our knowledge, this is the first time this approach has been adopted for group-feature selection. We now briefly describe the datasets employed in our study for feature selection.

 In the Iris data
set, the four features are sepal length ($f_1$), sepal
width ($f_2$), petal length ($f_3$) and petal width ($f_4$).

The  Wisconsin Breast Cancer (WBC) dataset records the measurements for 699 breast cancer cases. It has has nine features and two classes, benign and malignant. 

The Thyroid dataset comprises information from five laboratory tests conducted on a sample of 215 patients. The primary objective is to predict the classification of a patient's thyroid into one of three categories: euthyroidism, hypothyroidism, or hyperthyroidism.

The Sonar dataset has 208 observations, each representing the response of a sonar signal bounced off a metal cylinder or cylindrical rock at various angles. The sonar signals were collected using a single-beam echo sounder and are represented by 60 numerical attributes, which are the energy values within a specific frequency band. The dataset consists of 111 observations of rocks and 97 observations of mines, making it a reasonably balanced dataset.

The SRBCT dataset contains 83 samples, where each sample is characterized by 2,308 gene expression values. This dataset encompasses four distinct classes: 29 samples belong to the Ewing sarcoma (EWS), 18 to neuroblastomas (NB), 25 to rhabdomyosarcomas (RMS), and 11 to Burkitt lymphomas (BL).

The Leukemia dataset contains gene expressions of acute lymphoblastic leukemia (ALL) and acute myeloid leukemia (AML) samples originating from both bone marrow and peripheral blood. In total, the dataset consists of 72 samples, with 47 samples representing ALL and 25 samples representing AML. Each sample is characterized by expression values of 5147 genes.

\begin{figure*}[ht]
\centering
\centering
\subfloat[]{\includegraphics[width=0.48\linewidth,height=0.35\linewidth]{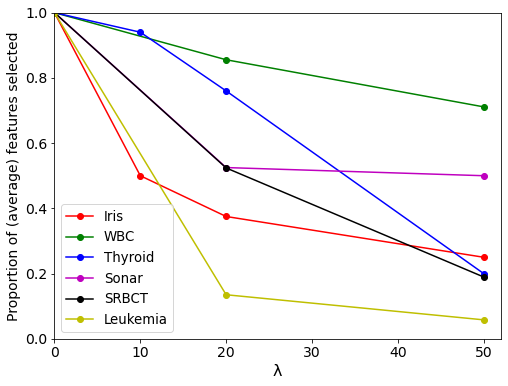}} 
\subfloat[]{\includegraphics[width=0.48\linewidth,height=0.35\linewidth]{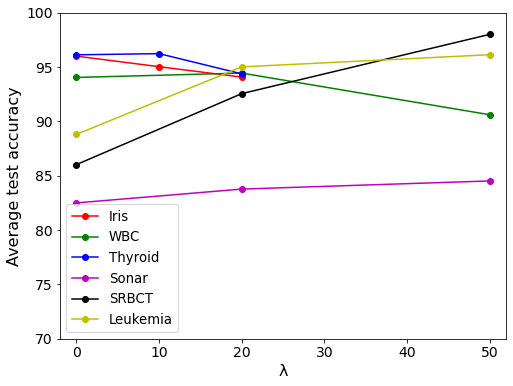}}
\caption[]{{  (a) Variation in the proportion of selected features  with the increase in  $\lambda$, the penalty factor
for redundancy, for six data sets.  (b)  Variation in the  average classification accuracy with the increase in $\lambda$ for six data sets.}}
\label{fig:lambda}  
\end{figure*}

Table \ref{tab:fsresult} summarizes the performance of the proposed method, considering only the penalty for feature redundancy.
It is evident from Table \ref{tab:fsresult} that as we increase the value of $\lambda$, the penalty for redundancy increases and so a lesser number of features are selected, and also the maximum absolute correlation among the selected features is reduced. However, the average absolute correlation  sometimes is slightly increased when the penalty is too high, i.e., when a very small number of features are selected, the maximum absolute correlation generally decreases, but not the average. It is worth noting that for Sonar, SRBCT, and Leukemia, the test accuracy improved with a reduced number of features on average. This is suggestive of the fact that feature selection with controlled redundancy can not only reduce the measurement and system design cost  by selecting a small set of useful features but also
can improve the accuracy. {  To offer a clearer insight into these dynamics, we have plotted the relationship between $\lambda$ and the average test accuracy as well as $\lambda$ and the proportion of features selected on average in Fig. \ref{fig:lambda}. Figure \ref{fig:lambda} has two panels (a) and (b).
Figure 1(a) depicts the impact of increasing $\lambda$ on the proportion of features selected by the proposed method. As expected, with an increase in $\lambda$ for all six datasets on average the proportion of selected features decreases. Figure 1(b), on the other hand, depicts the effect of increased $\lambda$ on the test accuracy. In general, with an increase in $\lambda$ less number of features gets selected which is likely to have a derogatory effect on the test accuracy. However, as already indicated that for high-dimensional data like SRBCT and Leukemia the test accuracy increases with the increase in $\lambda$.  A possible reason for this maybe that a high-dimensional data set is likely to have more redundancy in the feature set and hence with an increase in $\lambda$ at the beginning, the test performance improves. } 

\begin{figure}
    \centering
    \includegraphics[scale=0.5]{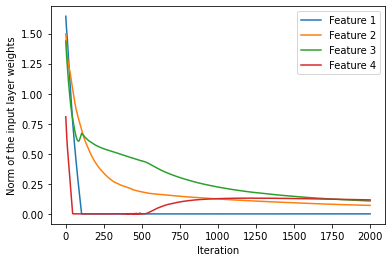}
    \caption{Variation of the norm of the weight vectors connecting input nodes to hidden nodes for the IRIS
data set with iterations for $\lambda=5$ and $\mu=0$.}
    \label{fig:iris}
\end{figure}

 Fig. \ref{fig:iris} shows the norms of the weight vectors connecting input nodes to hidden nodes of a typical run for the Iris data with $\lambda=5$ and $\mu=0$. 
We observe that the norm of the weight vector connecting the feature, Sepal Length (feature 1), to the hidden layer nodes, readily decreases with iteration and becomes close to zero. It is well known that Petal Length (feature 3) and
Petal Width (feature 4) are the most discriminatory features. Sepal Length is highly correlated with both of them (see Table \ref{tab:corr-iris}). Hence, the weight norm of Sepal Length is penalized to be very close to zero, whereas Sepal Width (feature 2), having a low correlation with all other features, has a higher weight norm. This behavior demonstrates the effectiveness of our penalty term in controlling redundancy and facilitating the selection of discriminative features. We emphasize here that unlike \cite{wang2020feature}, we do not use the group lasso penalty (eq. \eqref{gl}), which explicitly helps to reduce the number of selected features. Here, our goal is to demonstrate the effectiveness of the penalty for redundancy, that we have proposed.

\begin{table}[!htb]
    \centering
    \caption{Correlation matrix for IRIS data}
\label{tab:corr-iris}
    \resizebox{0.4\linewidth}{!}{
    \begin{tabular}{c|cccc}
    \toprule
    \addlinespace
    Feature & 1 &2 &3 &4\\
    \midrule
\addlinespace 
1 & 1.00   &  -0.11  &    0.87  &   0.82\\
2 & -0.11  &    1.00  &    -0.42   &  -0.36\\
3 & 0.87   &  -0.42   &   1.00   &  0.96\\
4 & 0.82  & -0.36    & 0.96  &   1.00\\
\bottomrule
 \end{tabular}}
 \end{table}

\begin{table*}[!htb]
    \centering
    \caption{Average test accuracy of the proposed method, RCFS, mFSMLP-Cor, and SGLC (Results of RCFS, mFSMLP-Cor, and SGLC are directly collected from \cite{wang2020feature})}
    \label{tab:comp-fs}
     \resizebox{\linewidth}{!}{
    \begin{tabular}{l|cc|cc|ccc|ccc}
\toprule
    \addlinespace
    \multirow[b]{2}{*}{ Datasets } & \multicolumn{2}{c}{ RCFS } & \multicolumn{2}{c}{ mFSMLP-CoR } & \multicolumn{3}{c}{ SGLC } & \multicolumn{3}{c}{ Proposed Method }\\
     %\multirow[c]{3}{*}{ Datasets } & \multicolumn[c]{2}{*}{ RCFS } & \multicolumn[c]{2}{*}{ mFSMLP-CoR } & \multicolumn[c]{3}{*}{ SGLC } & \multicolumn[c]{3}{*}{ Proposed Method }\\\\
% \cmidrule { 2 - 11 }& 
% Test Acc & 
% $\begin{array}{c}\text{Max- } \\\text{abs corr }\end{array}$ & Test Acc & $\begin{array}{c}\text{Max- } \\
% \text {abs corr }\end{array}$ & Test Acc & $\begin{array}{c}\text{Avg- } \\
% \text{abs corr }\end{array}$ & $\begin{array}{c}\text{Max } \\
% \text{abs corr }\end{array}$ & Test Acc & $\begin{array}{c}\text{Avg- } \\
% \text{abs corr }\end{array}$ & $\begin{array}{c}\text{Max } \\
% \text {abs corr }\end{array}$\\
\cmidrule { 2 - 11 } &
\specialcell{\textbf{Test}\\\textbf{Acc}} &
\specialcell{\textbf{Max abs}\\\textbf{corr}} &
\specialcell{\textbf{Test}\\\textbf{Acc}} &
\specialcell{\textbf{Max abs}\\\textbf{corr}} &
\specialcell{\textbf{Test}\\\textbf{Acc}} &
 \specialcell{\textbf{Average}\\\textbf{abs corr}} &
\specialcell{\textbf{Max abs}\\\textbf{corr}} &
\specialcell{\textbf{Test}\\\textbf{Acc}} &
\specialcell{\textbf{Average}\\\textbf{abs corr}} &
\specialcell{\textbf{Max abs}\\\textbf{corr}} \\

\midrule
\addlinespace 
Iris & 95.4 & 0.96 & 94.8 &  \textbf{0.42}   &  96.0   &  \textbf{0.42}   & 0.96 & \textbf{96.1}
& \textbf{0.42} &	\textbf{0.42}\\

Thyroid & 87.9 & 0.72 & 90.7 & 0.42 & 94.4 & \textbf{0.41} & \textbf{0.41} & \textbf{94.5} & 0.43 & 0.43\\

WBC & 94.6 & 0.90 & 95.3 & 0.69 &  96.0   &  0.64   & 0.69 & 95.4 & \textbf{0.59}
& \textbf{0.59}\\

Sonar & 75.4 & 0.91 & 77.6 & 0.63 &  78.4   &  0.62   & 0.79 & 77.5 & 	\textbf{0.20} & \textbf{0.62}\\

SRBCT & 92.0 & 0.97 & 95.0 & 0.68 &  95.3   &  0.59  & 0.76 & \textbf{95.7} & \textbf{0.21}
 & \textbf{0.64}
 \\
\bottomrule
\end{tabular}}
\end{table*}

\begin{table*}[!htb]
    \centering
    \caption{Group-feature Selection Result\label{tab:ssres}}
    \resizebox{\linewidth}{!}{
    \begin{tabular}{cc|cccccc|cccccc}
    \toprule
    \addlinespace
        $\bm{\lambda}$ &
       $\bm{\mu}$ &
        \textbf{Dataset} & 
        \specialcell{\textbf{Test}\\\textbf{Acc}} &
       \specialcell{\textbf{Distinct}\\\textbf{\# sensors}} &
        \specialcell{\textbf{Average}\\\textbf{\# sensors}} &
\specialcell{\textbf{Max}\\\textbf{dep}} &
\specialcell{\textbf{Avg}\\\textbf{dep}} &
\textbf{Dataset} &
\specialcell{\textbf{Test}\\\textbf{Acc}} &
\specialcell{\textbf{Distinct}\\\textbf{\# sensors}} &
       \specialcell{\textbf{Average}\\\textbf{\# sensors}} &
\specialcell{\textbf{Max}\\\textbf{dep}} &
\specialcell{\textbf{Avg}\\\textbf{dep}}  \\ 
\midrule
        \addlinespace
        0 & 0 & Iris & 96.30  & 2 & 2 & 0.71 & 0.59 & Iris 2 & 96.00  & 3 & 3 & 0.99 & 0.75 \\ 
        0 & 2 & ~ & 95.79  & 1 & 1 &0&0& ~ & 95.80  & 1 & 1 &0&0\\ 
        0 & 5 & ~ & 96.00  & 1 & 1 &0&0& ~ & 94.77  & 1 & 1 &0&0\\ 
        10 & 0 & ~ & 95.67  & 1 & 1 &0&0& ~ & 94.13  & 1 & 1 &0&0\\ 
        10 & 2 & ~ & 95.67  & 1 & 1 &0&0& ~ & 94.50  & 1 & 1 &0&0\\ 
        10 & 5 & ~ & 95.30  & 1 & 1 &0&0& ~ & 96.19  & 1 & 1 &0&0\\ 
        20 & 0 & ~ & 95.70  & 1 & 1 &0&0& ~ & 95.40  & 1 & 1 &0&0\\ 
        20 & 2 & ~ & 96.00  & 1 & 1 &0&0& ~ & 94.73  & 1 & 1 &0&0\\ 
        20 & 5 & ~ & 96.67  & 1 & 1 &0&0& ~ & 94.26  & 1 & 1 &0&0\\ 
        \addlinespace
        \hline
        \addlinespace
        0 & 0 & Smart & 92.54  & 16 & 16 & 0.99 & 0.52 & Gas  & 98.40  & 16 & 16 & 0.99 & 0.52\\ 
        0 & 2 & Phone & 92.64  & 7 & 6.1 & 0.97 & 0.46 & Sensor & 97.61  & 7 & 5.5 & 0.97 & 0.46 \\ 
        0 & 5 & ~ & 92.57  & 6 & 5.6 & 0.97 & 0.46 &  & 96.23  & 5 & 4.3 & 0.84 & 0.43 \\ 
        20 & 0 & ~ & 91.21  & 7 & 5.6 & 0.97 & 0.50  & ~ & 97.98  & 7 & 6.8 & 0.97& 0.44 \\ 
        20 & 2 & ~ & 91.27  & 7 & 6.8 & 0.97 & 0.50  & ~ & 97.68  & 7 & 5.5 & 0.97& 0.44 \\ 
        20 & 5 & ~ & 92.94  & 7 & 7 & 0.97 & 0.46 & ~ & 94.15  & 5 & 3.5 & 0.68 & 0.40  \\ 
        50 & 0 & ~ & 89.40  & 5 & 4.5 & 0.95 & 0.52 & ~ & 97.76  & 6 & 5.8 & 0.97 & 0.43\\ 
        50 & 2 & ~ & 91.36  & 7 & 6.1 &0.97 & 0.50 & ~ & 97.66  & 6 & 5.5 & 0.96 & 0.44 \\ 
        50 & 5 & ~ & 91.82  & 6 & 6 & 0.97 & 0.50& ~ & 90.84  & 4 & 2.7 & 0.72 & 0.44 \\ 
        \addlinespace
        \hline
        \addlinespace
        0 & 0 & LRS & 82.86 & 2 & 2 & 0.78 & 0.75 & LandSat & 84.47 & 4 & 4 & 0.88 & 0.82 \\ 
        0 & 2 & ~ & 82.73 & 2 & 1.6 & 0.47 & 0.45 & ~ &  84.53 & 4 & 4 & 0.88 & 0.82 \\
        0 & 5 & ~ & 83.75 & 2 & 1.4 & 0.31 & 0.30 & ~ & 84.60 & 4 & 4 & 0.88 & 0.82 \\
        20 & 0 & ~ & 83.27 & 2 & 1.5 & 0.39 & 0.38 & ~ & 84.54 & 4 & 4 & 0.88 & 0.82 \\ 
        20 & 2 & ~ & 84.11 & 2 & 1.3 & 0.23 & 0.22 & ~ & 84.47 & 4 & 4 & 0.88 & 0.82 \\
        20 & 5 & ~ & 83.60 & 2 & 1.2 & 0.16 & 0.15 & ~ & 84.54 & 4 & 4 & 0.88 & 0.82 \\
        50 & 0 & ~ & 83.29 & 2 & 1.2 & 0.16 & 0.15 & ~ & 84.57 & 4 & 4 & 0.88 & 0.82 \\ 
        50 & 2 & ~ & 83.56 & 2 & 1.1 & 0.08 & 0.07 & ~ & 84.50 & 4 & 4 & 0.88 & 0.82 \\
        50 & 5 & ~ & 84.06 & 1 & 1 &0&0& ~ & 84.54 & 4 & 4 & 0.88 & 0.82 \\
        \bottomrule
\end{tabular}}
\end{table*}

 \subsubsection{Comparison with existing neural-network-based methods}

In this comparative analysis, our proposed method is evaluated alongside three state-of-the-art neural-network-based methods with redundancy handling ability: Redundancy-Constrained Feature Selection (RCFS) \cite{zhou2010feature}, modified Feature Selection using Multilayer Perceptron with Control on Redundancy (mFSMLP-CoR) \cite{chakraborty2014feature}, and Sparse Group Lasso with Control (SGLC) \cite{wang2020feature}. RCFS performs feature selection based on trace-based class separability while constraining redundancy. It provides the flexibility for users to specify the desired number of features. On the other hand, mFSMLP-CoR represents an enhanced version of FSMLP-CoR, featuring improved learning processes.
 In Table \ref{tab:comp-fs}, using different data sets, we compare the performance of our method with that of  RCFS, mFSMLPCoR, and SGLC. For  RCFS, mFSMLP-CoR, and SGLC the results are directly obtained from \cite{wang2020feature}. We use the cross-validation method to determine the number of hidden nodes and the parameters $\lambda$.  
To make a fair comparison, in this table, for all three methods, we use the same number of selected features. The best results, i.e, the maximum value of the test accuracy and the minimum value of 
the average and the maximum absolute correlation among the selected features are boldfaced. 
 %The mFSMLP-CoR reported the maximum absolute pairwise correlations over the tenfold experiment. However, the global maximum over the tenfold may be misleading. For example, in nine of the ten folds, the maximum correlation could be very low and in just one fold; it could be high resulting in the value of the max-correlation very high. This may give a false perception of the results. Hence, along with the maximum correlation, we also report the average of the maximum absolute correlations over the tenfold. 
 We can see from Table \ref{tab:comp-fs} that for majority of the datasets, our method outperforms the state-of-the-art methods, both in terms of the capability of reducing the maximum or the average absolute correlation among the selected features and the test accuracy. These results underscore the efficacy of our approach in feature selection with controlled redundancy, positioning it as a promising solution compared to existing methods.

\subsubsection{Comparison with existing non-neural-network-based methods}
\label{non-nn-expt}
\begin{figure*}[ht]
\centering
\centering
\subfloat[Lung dataset]{\includegraphics[width=0.48\linewidth]{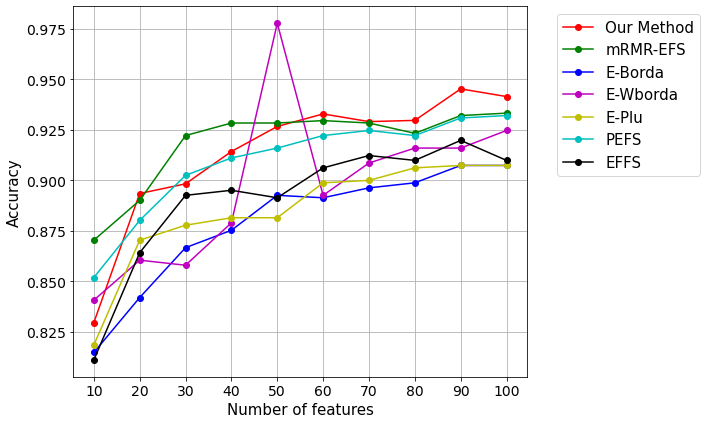}} 
\subfloat[ DLBCL dataset]{\includegraphics[width=0.48\linewidth]{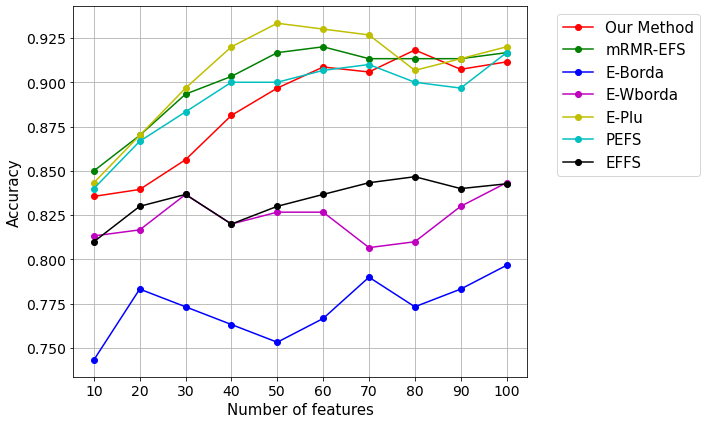}}\\
\subfloat[NCI60 dataset]{\includegraphics[width=0.48\linewidth]{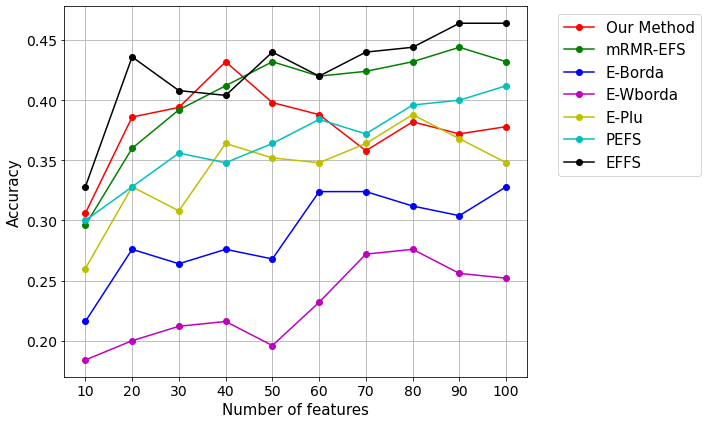}} 
\subfloat[Semeion dataset]{\includegraphics[width=0.48\linewidth]{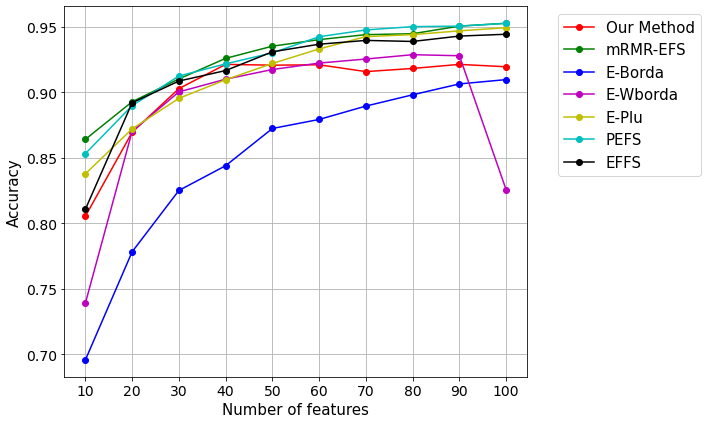}}
\caption[]{ Variation of average accuracy is plotted with the number of selected features}
\label{fig:comp-efs}  
\end{figure*}

In this section, we compare our method to
the six state-of-the-art ensemble FS methods based on rank aggregation.
These methods are E-Borda \cite{drotar2019ensemble} which uses the Borda-count
method, E-WBorda \cite{drotar2019ensemble} uses weighted Borda-count, E-Plu \cite{drotar2019ensemble} uses plurality voting, and PEFS \cite{hashemi2021pareto}, EFFS \cite{tian2020robust} use
the mean and minimum rank assigned to each feature
by feature selection methods and mRMR-EFS \cite{hashemi2023minimum}.  The results for these methods were directly obtained from \cite{hashemi2023minimum}. To ensure a fair comparison, we adhered to the same setup described in \cite{hashemi2021pareto}: $60\%$ of the samples are used as the training data, with the remaining $40\% $  reserved for testing. The average accuracies obtained from 20 separate runs for each method are reported. We performed 10-fold cross-validation to select the number of hidden nodes $h$ and the parameters $\lambda,\mu$. We performed the cross-validation on a grid with $h=5,6,\cdots,20$, $\mu=10,20,30,\cdots,100$, and $\lambda=5,10,15,\cdots,50$. The average accuracy for different number of features is illustrated in Fig \ref{fig:comp-efs}. From the graphs in Fig \ref{fig:comp-efs}, it is evident that no single method consistently outperforms the others. 

To summarize the results, we introduce a scoring or ranking scheme: we define the score of a method as the number of values on the x-axis (i.e., the number of features selected) for which that method outperforms all other methods. For example, in the case of the lung dataset, our method outperforms all other methods when the numbers of features selected are 20, 60, 70, 80, 90, and 100, resulting in a score of 6. Similarly, mRMR-EFS has a score of 3, E-WBorda has a score of 1, and all other methods have a score of 0. Therefore, for the lung dataset, our method ranks 1st, mRMR-EFS ranks 2nd, E-WBorda ranks 3rd, and all other methods rank 4th. According to our ranking scheme, our method performs the best for the lung dataset. We applied this ranking process across all datasets and calculated the average rank. It turns out that both our method and mRMR-EFS have the highest average rank of $2.25$.

\subsection{Group-feature (sensor) selection results}
\label{subsec:ss-res}

Table \ref{tab:ssres}  summarizes the results on a set of data sets for group-feature selection
using different choices of the parameters $\lambda$, and $\mu$. We now briefly describe the datasets and the groups of features used in this study.

In the Iris dataset, we consider two natural groups: the first group includes sepal length and width, while the second group comprises the remaining two features.

The second version of the Iris dataset, denoted as Iris 2, incorporates seven features: $f_1, f_2, f_3, f_4$, $f_5 = f_1 + N(0, 0.05), f_6 = f_3 + N(0, 0.05)$ and
$f_7 = f_4 + N(0, 0.05)$. Thus, for this dataset, the pairs ($f_1, f_5$), ($f_3, f_6$) and ($f_4, f_7$) are strongly correlated. We group these seven features into three groups. The first group comprises $f_1$ and $f_2$, the second group includes $f_3, f_4$, and $f_5$, and the last group contains $f_6$ and $f_7$.

The Smartphones dataset is a six-class data, where each class represents one of the six activities performed by a subject while carrying a smartphone. The six activities/classes are: walking, walking upstairs, walking downstairs, sitting, standing, and lying. The data set includes recordings of 17 signals of 30 subjects. From these 17 signals, features like correlation, or autoregressive coefficients were extracted. This resulted in 17 groups of features with different numbers of features in different groups.

The Gas Sensor dataset is a 128-dimensional dataset that represents recordings from 16 chemical sensors that are exposed to six gases at different concentration levels. Thus, this is a six-class dataset and the task is to discriminate among the six different gases.

The low-resolution spectrometer (LRS) dataset contains 531 spectra derived from IRAS-LRS database. It has 44 features from the blue band and 49 features from the red band (total 93 features).  
%features from two bands - blue and red. These two bands consist of 44 and 49 flux measurements, respectively, making it a 93-dimensional dataset with two groups/sensors.

On the other hand, the LandSat dataset is a 6-class Statlog (Landsat Satellite) data. 
There are four spectral bands. For each pixel is represented by 9 intensity values taken from a $3\times 3$ neighborhood for that pixel. This is done for all four bands.
There are six classes, and the class label corresponds to the center pixel of the $3 \times 3$ block. As done in \cite{chakraborty2014sensor}, we have used the mean and standard deviation computed over a $3\times 3$ window and augmented these with the other features. 
Consequently, the dataset used in our study, referred to as LandSat, consists of four groups, each with 11 (9+2) features.
There are a total of 6435 sample points distributed among the six classes. For our experiments, we utilized 4435 samples for the training set and reserved 2000 samples for testing, following the recommendations from the UCI repository.

Table \ref{tab:ssres} summarizes the results on these datasets. When only one sensor (group-feature) was selected, we defined the maximum and average dependency of the selected sensor to be zero. This is reasonable since if only one sensor is selected, there is NO dependency in the set of selected groups and redundancy is minimized. It is evident that as we increase the value of $\lambda$ or $\mu$, the penalty for redundancy or the group lasso penalty increases. Consequently, a fewer groups of features should be selected and also the maximum and average dependency among the selected groups should decrease. Table \ref{tab:ssres} demonstrates that, except for LandSat, this trend generally holds across the datasets. But, notably, the impact of the reduced number of features due to higher values of the regularizers on the average accuracy is not much. In fact, in some cases, the test accuracy increases with higher penalties. For instance, in the Iris dataset, the test accuracy marginally improves with the most severe penalty ($\lambda=20,\mu=5$), similar to the trend observed in the LRS dataset.

\begin{figure}[!htb]
    \centering
    \includegraphics[scale=0.57]{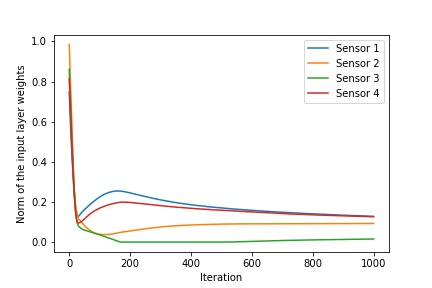}
    \caption{Variation of the norm of the weight vectors connecting input nodes to hidden nodes for  the LandSat
data set with iterations for $\lambda=20$ and $\mu=1$.}
    \label{fig:landsat}
\end{figure}

For LandSat, we have observed from Table \ref{tab:ssres} that our method is unable to reduce the number of sensors with the 
threshold $\theta=0.1\times \max_i \|\mathbf{g}_i\|$. However, from
Fig. \ref{fig:landsat}, we can see that the norms of the weights of sensors 1 and 4 are close to each other and are notably higher than those of sensors 2 and 3. From Table \ref{tab:dep-landsat}, we observe that sensors 1 and 4 have the least level of dependency between them, whereas sensors 2 and 3 are highly dependent on all other sensors. These findings (Fig. \ref{fig:landsat}) align with our intended objective. 
In line with this, rather than employing a fixed threshold, we opt to select the top two sensors in this specific scenario.  This is what we do when we compare with the existing method(s) - this philosophy also ensures a fair comparison with the results in \cite{chakraborty2014sensor} (see Table \ref{tab:comp_ss}).

\begin{table}[!htb]
    \centering
    \caption{Group dependency matrix for the LandSat data}
\label{tab:dep-landsat}
    \resizebox{0.4\linewidth}{!}{
    \begin{tabular}{c|cccc}
    \toprule
    \addlinespace
    Group & 1 &2 &3 &4\\
    \midrule
\addlinespace 
1 & 1.00   &  0.87 &  0.81 &  0.69\\
2 & 0.88 &  1.00 & 0.88 & 0.79\\
3 & 0.79 & 0.88 &  1.00 &  0.88\\
4 & 0.68 &  0.81 &  0.87 &  1.00\\
\bottomrule
 \end{tabular}}
 \end{table}
 
\subsubsection{Comparison with existing method}

We now compare our GFS results with mGFSMLP-CoR proposed by Chakraborty and Pal \cite{chakraborty2014sensor}. The results in Table \ref{tab:comp_ss} are attained considering the same number of selected features for both methods. This ensures a fair comparison. Our implementation follows Algorithm 1, with a slight deviation in line 12. Instead of using a fixed threshold, we select the \emph{same} number of sensors as the rounded average number of sensors from Table 18 of Chakraborty and Pal  \cite{chakraborty2014sensor}, to obtain the reduced data. We have reported the test accuracy for both methods in Table \ref{tab:comp_ss}, with the best outcomes highlighted in bold for clarity. We performed 10-fold cross-validation to select the number of hidden nodes $h$ and the parameters $\lambda$ and $\mu$.

\begin{table}[!htb]
    \centering
    \caption{Average test accuracy of the proposed method and mGFSMLP-CoR (Results of mGFSMLP-CoR are directly collected from \cite{chakraborty2014sensor})}
    \label{tab:comp_ss}
    \resizebox{0.6\linewidth}{!}{
    \begin{tabular}{lcc}
    \toprule
    \addlinespace
        \textbf{Dataset} & mGFSMLP-CoR & Proposed Method\\
    \midrule \addlinespace
         Iris & 96.07 & \textbf{96.64}
\\
         Iris 2 & 96.07 & \textbf{96.58}
 \\
         LRS &  \textbf{88.68} & 83.86
\\
         Gas Sensor & 88.80 & \textbf{97.23}
 \\
 LandSat & 81.65 & \textbf{84.60}\\
    \bottomrule
    \end{tabular}}
    
\end{table}

Our method outperforms mGFSMLP-CoR \cite{chakraborty2014sensor} in the majority of datasets (in four out of five datasets). Unfortunately, we could not implement our method on the other five datasets reported in \cite{chakraborty2014sensor}, due to the unavailability of those datasets online. Nonetheless, the results obtained on the datasets where our method was applied, demonstrate its effectiveness in group-feature (sensor) selection with controlled redundancy.

\subsection{Application on RNA Sequence dataset}
\label{subsec-rna}
We conclude our experimental section by demonstrating an application of our feature selection and group-feature selection methods on a gene expression cancer RNA-Seq dataset \cite{misc_gene_expression_cancer_rna-seq_401}.

In the RNA-Seq dataset, for 801 patients with five different kinds of tumors (BRCA, KIRC, COAD, LUAD and PRAD) the expression levels of 20,531 genes were measured by the Illumina HiSeq platform. We considered two cases: When each group consists of one gene (denoted RNA Seq 1) and when (after removing the last feature) every five consecutive genes were grouped into one group (denoted RNA Seq 2), as done in Tang et al. \cite{tang2018group}. 

\begin{figure*}[ht]
\centering
\centering
\subfloat[]{\includegraphics[width=0.48\linewidth,height=0.35\linewidth]{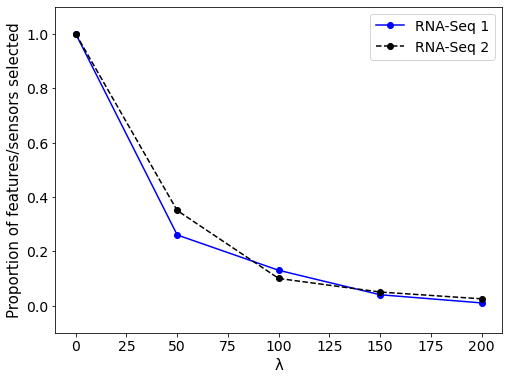}} 
\subfloat[]{\includegraphics[width=0.48\linewidth,height=0.35\linewidth]{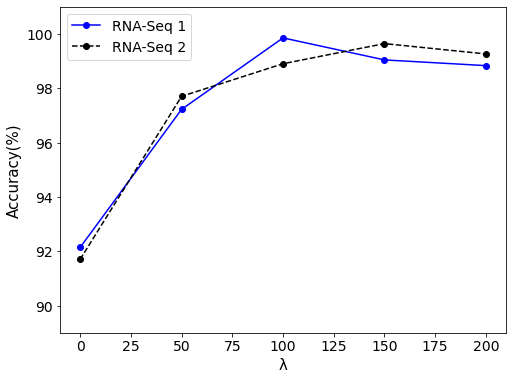}}
\caption[]{ { (a) Variation of the proportions of selected features with the penalty parameter $\lambda$ for RNA-Seq 1 and RNA-Seq 2.  (b) Variation of accuracy  with the penalty parameter $\lambda$ for RNA-Seq 1 and RNA-Seq 2. }}
\label{fig:lambda-rna}  
\end{figure*}

\begin{figure*}[ht]
\centering
\centering
\subfloat[RNA-Seq 1]{\includegraphics[width=0.48\linewidth,height=0.35\linewidth]{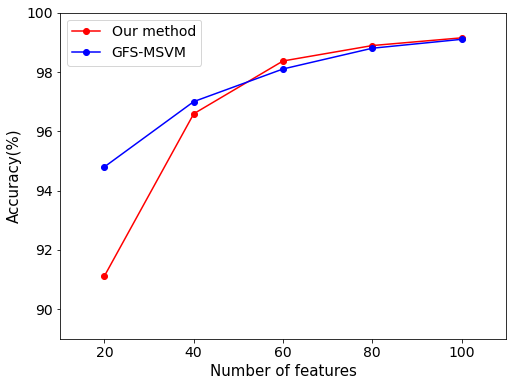}} 
\subfloat[RNA-Seq 2]{\includegraphics[width=0.48\linewidth,height=0.35\linewidth]{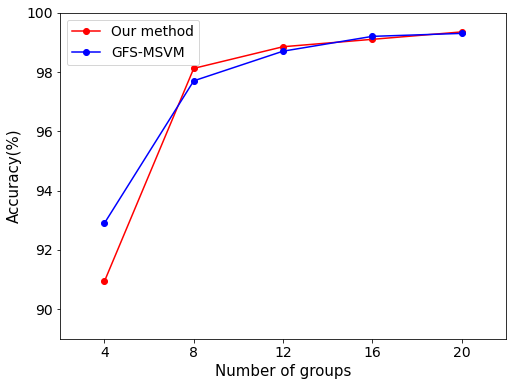}}
\caption[]{{ (a) Changes in accuracy with increase in the number of selected features for RNA-Seq 1. (b) Changes in accuracy with increase in the number of selected groups of features for RNA-Seq 2. Both the GFS-MSVM and our method demonstrate good accuracy. Specifically, our method shows a slight improvement over GFS-MSVM when a larger number of features/sensors are selected.}}
\label{fig:rna-comp}  
\end{figure*}

We conducted experiments with five different values of $\lambda$. The changes in the proportion of selected features/sensors and the corresponding accuracy with varying $\lambda$ are illustrated in Figure \ref{fig:lambda-rna}. Our observations indicate that a smaller number of features/sensors are selected as the regularization parameter increases. Additionally, we note that the accuracy achieved with the selected features/sensors surpasses that obtained with all features, thus underscoring the effectiveness of our method.

\subsubsection{Comparison with GFS-MSVM \cite{tang2018group}}
\label{comp-rna} 

We compare the performance of our methods with the GFS-MSVM (Group Feature Selection for Multiclass Support Vector Machines) method proposed by Tang et al. \cite{tang2018group}. Our implementation follows Algorithm 1, with slight deviations in lines 2 and 12 to match the settings of  Tang et al. \cite{tang2018group}. Instead of using an 80\%-20\% sample-splitting strategy, we select 20 samples for each class as training samples, resulting in 100 training samples, while the remaining samples are used for testing. Additionally, instead of employing a fixed threshold, we choose a pre-fixed number (denoted as $d$) of features/sensors (having the top $d$ values of the weights) to obtain the reduced data. The accuracy of both methods for different values of the number of selected features is shown in Figure \ref{fig:rna-comp}. Here, our method exhibits slightly better accuracy when a larger number of features are selected.

\begin{comment}

\begin{table*}[!htb]
    \centering
    \caption{Average test accuracy of the proposed method and GFS-MSVM \cite{tang2018group}}
    \label{tab:comp_rna}
     \resizebox{0.9\linewidth}{!}{
    \begin{tabular}{l|cc|cc}
\toprule
    \addlinespace
    \multirow[b]{2}{*}{ Number of features } & \multicolumn{2}{c}{ RNA Seq 1 } & \multicolumn{2}{c}{ RNA Seq 2 } \\

\cmidrule { 2 - 5 } &
GFS-MSVM &
Proposed Method &
GFS-MSVM &
Proposed Method  \\

\midrule
\addlinespace 
20 feature/ 4 groups & 95.4 & 0.96 & 94.8 &  \textbf{0.42}  \\

40 feature/ 8 groups & 87.9 & 0.72 & 90.7 & 0.42 \\

60 feature/ 12 groups & 94.6 & 0.90 & 95.3 & 0.69 \\

80 feature/ 16 groups & 75.4 & 0.91 & 77.6 & 0.63 \\

100 feature/ 20 groups & 92.0 & 0.97 & 95.0 & 0.68 
 \\
\bottomrule
\end{tabular}}
\end{table*}
\end{comment}

\section{Conclusion}
\label{sec:conc}
This article has introduced an integrated feature selection scheme, that effectively controls the degree of redundancy in the set of selected features, and we have further generalized it for sensor (group-feature) selection. We have also generalized the group lasso regularization and incorporated it alongside the penalty for controlling redundancy for sensor selection, offering a unified method for the selection of feature groups (sensors), which is capable of efficiently eliminating uninformative groups, preserving valuable ones, and keeping control over the level of redundancy. Our proposed regularizers for controlling redundancy in both feature and sensor selection offer an effective and more intuitive approach compared to existing methods. The experimental results on different commonly used datasets demonstrated the efficacy of our method. The proposed model achieved competitive performance in terms of classification accuracy while significantly reducing the number of selected features/sensors compared to existing methods. This reduction in feature dimensionality not only improves computational efficiency but may also enhance interpretability and reduce the risk of overfitting.
In this work, the choice of parameters ($\mu \text{ and } \lambda$) has been made in an ad hoc manner, but systematic methods, such as cross-validation can also be used to select these parameters if necessary. Furthermore, the underlying philosophy can be readily extended to other machine learning models, including radial basis function (RBF) networks, offering a versatile tool for feature selection and redundancy control in various applications and domains in the field of machine learning.

\section*{Acknowledgments}
We thank the reviewers for their careful reading of an earlier version of the article. Their insightful comments and constructive feedback have significantly contributed to the refinement and improvement of our work.




\bibliographystyle{splncs04}

\end{document}